\documentclass[11pt,a4paper]{article}
\usepackage[final]{emnlp2021}
\usepackage{times}
\usepackage{latexsym}
\usepackage{todonotes}
\usepackage{subcaption}
\usepackage{xspace}
\usepackage{amsmath}
\usepackage{amssymb}
\usepackage{bbm}
\usepackage{multirow}
\usepackage{booktabs}

\usepackage{microtype}

\def\aclpaperid{3574} %

\newcommand{\ie}{i.e.,\xspace}
\newcommand{\eg}{e.g.,\xspace}
\newcommand{\eat}[1]{}

\title{Beyond Preserved Accuracy: \\ Evaluating Loyalty and Robustness of BERT Compression}

\author{Canwen Xu$^1$\thanks{\ \ Equal Contribution. Work done at Microsoft Research Asia.} , Wangchunshu Zhou$^{2*}$, Tao Ge$^3$, Ke Xu$^4$, Julian McAuley$^1$, Furu Wei$^3$\\
 $^1$ University of California, San Diego
 $^2$ Stanford University \\
 $^3$ Microsoft Research Asia
 $^4$ Beihang University \\
 $^1$\texttt{\{cxu,jmcauley\}@ucsd.edu,} $^2$\texttt{wcszhou@stanford.edu} \\
 $^3$\texttt{\{tage,fuwei\}@microsoft.com,}
 $^4$\texttt{kexu@nlsde.buaa.edu.cn} \\
}

\date{}

\begin{document}
\maketitle
\begin{abstract}
Recent studies on compression of pretrained language models (\eg BERT) usually use preserved accuracy as the metric for evaluation. In this paper, we propose two new metrics, label loyalty and probability loyalty that measure how closely a compressed model (\ie student) mimics the original model (\ie teacher). We also explore the effect of compression with regard to robustness under adversarial attacks. We benchmark quantization, pruning, knowledge distillation and progressive module replacing with loyalty and robustness. By combining multiple compression techniques, we provide a practical strategy to achieve better accuracy, loyalty and robustness.\footnote{Our code is available at \url{https://github.com/JetRunner/beyond-preserved-accuracy.}}
\end{abstract}

\section{Introduction}
Recently, many large pretrained language models (PLMs, \citealp{bert,roberta,xlnet,megatron,t5}) have been proposed for 
a variety of
Natural Language Processing (NLP) tasks. However, as pointed out in recent studies~\cite{strubell2019energy,greenai,parrot}, these models suffer from computational inefficiency and high ecological cost. Many attempts have been made to address this problem, including quantization~\cite{q8bert,qbert}, pruning~\cite{headprune,mvp}, knowledge distillation (KD)~\cite{distilbert,bertpkd,mobilebert,pd,tinybert,minilm,zhou2021meta} and progressive module replacing~\cite{bot}. 

\begin{figure}
    \centering
    \includegraphics[width=\columnwidth]{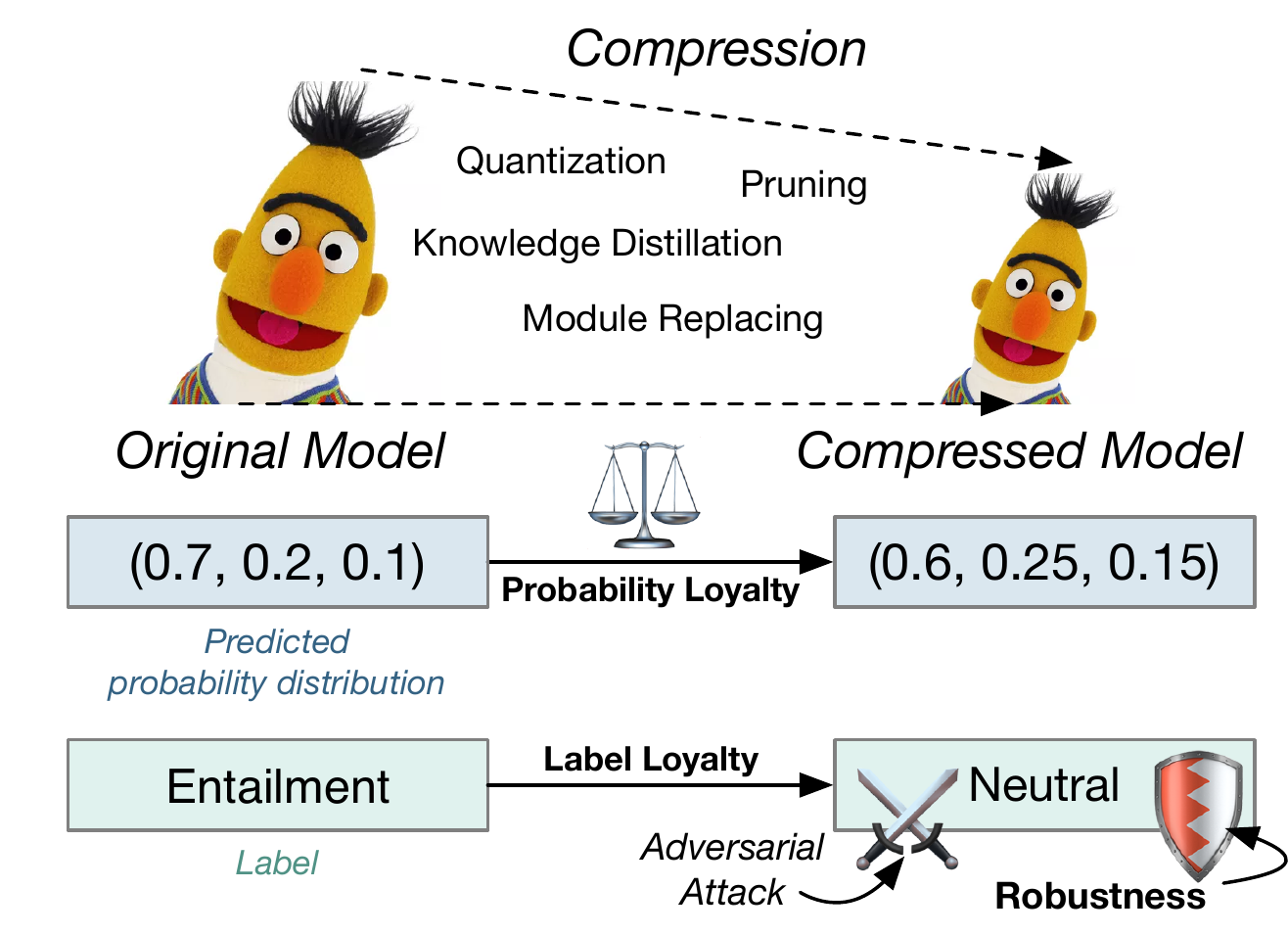}
    \caption{Three metrics to evaluate the compressed models beyond preserved accuracy. For each input, \textbf{label and probability loyalty} measure the shift of label and predicted probability distribution, respectively. \textbf{Robustness} measures the performance of the compressed model under adversarial attacks.}
    \label{fig:loyalty}
\end{figure}

BERT~\cite{bert} is a representative PLM.
Many works compressing BERT use preserved accuracy with computational complexity (\eg speed-up ratio, FLOPS, number of parameters) as metrics to evaluate 
compression. 
This evaluation scheme is far from perfect: (1) Preserved accuracy cannot reflect how alike the teacher and student\footnote{Teacher and student are originally concepts in knowledge distillation. In this paper, we will sometimes use teacher and student to refer to the original model and compressed model in other compression methods for simplicity.} models behave. This can be 
problematic when applying 
compression techniques 
in
production (to be detailed in Section \ref{sec:loyalty}). (2) Using preserved accuracy to evaluate models compressed with more data or data augmentation~\cite{tinybert} can be misleading, since one cannot tell whether the improvement should be attributed to the innovation of the compression technique or addition of data. (3) Model robustness, which is critical for production, is often missing from evaluation, leaving a possible safety risk.

As illustrated in Figure \ref{fig:loyalty}, to measure the resemblance between the student and teacher models, we propose \emph{label loyalty} and \emph{probability loyalty} 
to
target different but 
important aspects. 
We also explore the robustness of the compressed models by conducting black-box adversarial attacks. We apply representative BERT compression methods of different types to the same teacher model and benchmark their performance in terms of accuracy, speed, loyalty and robustness. We find that methods with a knowledge distillation loss perform well on loyalty and
that
post-training quantization can drastically improve 
robustness against 
adversarial attacks. We use the conclusions drawn from these experiments to combine multiple techniques together and achieve significant improvement in terms of accuracy, loyalty and robustness.

\section{BERT Compression}
Compressing and accelerating pretrained language models like BERT has been an active field of research. Some initial work employs conventional methods for neural network compression to compress BERT. For example, Q8-BERT~\citep{q8bert} and Q-BERT~\citep{qbert} employ weight quantization to reduce the number of bits used to represent 
a parameter
in a BERT model. Pruning methods like Head Prune~\citep{headprune} and Movement Pruning~\citep{mvp} remove weights based on their importance to reduce the memory footprint of pretrained models. Another line of research focuses on exploiting the knowledge encoded in a large pretrained model to improve the training of more compact models. For instance, DistilBERT~\citep{distilbert} and BERT-PKD~\citep{bertpkd} employ knowledge distillation~\citep{kd} to train compact BERT models in a task-specific and task-agnostic fashion respectively by mimicking the behavior of large teacher models. Recently, \citet{bot} proposed progressive module replacing, which trains a compact student model by progressively replacing the teacher layers with their more compact substitutes.

\begin{table*}[tb]
\begin{center}
\resizebox{1.\linewidth}{!}{
\begin{tabular}{l|cl|c|cc|cc}
\toprule

\multirow{2}{*}{Method} & \multirow{2}{*}{\#~Layer} & Speed & MNLI & \multicolumn{2}{c|}{Loyalty} & \multicolumn{2}{c}{Adversarial Attack} \\
& & -up $\uparrow$ & m/mm $\uparrow$ & Label $\uparrow$ & Probability $\uparrow$ & Acc $\uparrow$ & \#Query $\uparrow$ \\
\midrule
Teacher~\cite{bert} & 12 & 1.0$\times$ & $84.5$~/~$83.3$ & $100$ & $100$ & $8.1(\pm0.1)$ & $89.6(\pm0.1)$ \\
Truncate \& Finetune  & 6 & 2.0$\times$ & $81.1$~/~$80.0$ & $87.7(\pm0.2)$ & $84.9(\pm0.7)$ & $4.4(\pm0.1)$ & $78.0(\pm0.1)$\\
Pure KD & 6 & 2.0$\times$ & $81.1$~/~$80.8$ & $89.2(\pm0.1)$ & $89.5(\pm0.2)$ & $6.2(\pm0.1)$ & $80.1(\pm0.2)$  \\
\midrule
Q8-PTQ~\cite{q8bert} & 12 & 1.8$\times^\ddagger$ & $80.7$~/~$80.4$ & $89.6(\pm0.5)$ & $80.8(\pm0.4)$ & $40.2(\pm0.1)$ & $91.6(\pm0.1)$ \\
Q8-QAT$^\dagger$~\cite{q8bert} & 12 & 1.8$\times^\ddagger$ & $83.4$~/~$82.4$ & $89.7(\pm0.2)$ & $88.2(\pm0.3)$ & $6.8(\pm0.2)$ & $82.7(\pm0.2)$\\
\midrule
Head Prune~\cite{headprune} & 12 & 1.2$\times$ & $80.9$~/~$80.6$ & $87.8(\pm0.1)$ & $85.5(\pm0.6)$ & $9.1(\pm0.1)$ & $90.5(\pm0.2)$\\
\midrule
DistilBERT$^\dagger$~\cite{distilbert} & 6 & 2.0$\times$ & $82.4$~/~$81.4$ & $88.9(\pm0.2)$ & $88.4(\pm0.4)$  & $5.9(\pm0.1)$ & $80.8(\pm0.2)$\\
TinyBERT$^\dagger$~\cite{tinybert} & 6 & 2.0$\times$ & $82.7$~/~$82.7$ & $88.9(\pm0.1)$ & $88.4(\pm0.7)$ & $6.7(\pm0.1)$ & $82.1(\pm0.2)$ \\
\midrule
BERT-PKD~\cite{bertpkd} & 6 & 2.0$\times$ & $81.3$~/~$81.1$ & $88.9(\pm0.1)$ & $89.0(\pm0.2)$ & $6.4(\pm0.2)$ & $81.9(\pm0.2)$  \\
\midrule
BERT-of-Theseus~\cite{bot} & 6 & 2.0$\times$ & $81.8$~/~$80.7$ & $88.1(\pm0.2)$ & $82.5(\pm0.3)$ & $8.3(\pm0.2)$ & $89.7(\pm0.2)$\\

\bottomrule
\end{tabular}
}
\end{center}

\caption{Accuracy, loyalty and robustness of compressed models on the test set of MNLI (3 runs). 
Accuracy scores are from the GLUE~\cite{glue} test server. $^\dagger$These models are not initialized from (a part of) the finetuned BERT teacher. $^\ddagger$The speed-up ratio of quantization is benchmarked on CPU. $^\uparrow$Higher is better. }
\label{tab:main}
\end{table*}

\section{Metrics Beyond Accuracy}
\label{sec:loyalty}

\subsection{Loyalty}

\subsubsection{Label Loyalty}
Model compression is a common practice to optimize the efficiency of a model for deployment~\cite{cheng2017survey}. In real-world 
settings,
training and deployment are often separate~\cite{paleyes2020challenges}. 
As such
it is 
desirable
to have a metric to measure to what extent 
the ``production model'' is different from the ``development model''. Moreover, when discussing ethical concerns, previous studies \cite{minilm,pabee} 
ignore the risk that model compression could introduce additional biases.
However, a 
recent work~\cite{hooker2020characterising} strongly contradicts this assumption. In a nutshell, we would desire the student to behave as closely 
as possible to
the teacher, to make it more predictable and minimize the risk of introducing extra bias. Label loyalty directly reflects the resemblance of the labels predicted between the teacher and student models. It is calculated in the same way as accuracy, but between the student's prediction and the teacher's prediction, instead of ground labels:
\begin{equation}
    L_{l} = \mathrm{Accuracy}(\mathit{pred}_{t}, \mathit{pred}_{s})
\end{equation}
where $\mathit{pred}_{t}$ and $\mathit{pred}_{s}$ are the predictions of the teacher and student, respectively.

\subsubsection{Probability Loyalty}
Except for the label correspondence, we argue that the predicted probability distribution matters as well. In industrial applications, calibration~\cite{calibration,li2020accelerating}, which focuses on the meaningfulness of confidence, is an important issue for deployment.
Many dynamic inference acceleration methods~\cite{deebert,righttool,fastbert,xin2020early,li2020accelerating} use entropy or the maximum value of the predicted probability distribution as the signal for early exiting. Thus, a shift of predicted probability distribution in a compressed model could break the calibration and invalidate calibrated early exiting pipelines.

Kullback–Leibler (KL) divergence is often used to measure how one probability distribution is different from a reference 
distribution.
\begin{equation}
D_{\mathrm{KL}}(P \| Q)=\sum_{x \in \mathcal{X}} P(x) \log \left(\frac{P(x)}{Q(x)}\right)
\label{equ:kl}
\end{equation}
where $\mathcal{X}$ is the probability space; $P$ and $Q$ are predicted probability distributions of the teacher and students, respectively. Here, we use its variant, the Jensen–Shannon (JS) divergence, since it is symmetric and always has a finite value which is desirable for a distance-like metric:
\begin{equation}
D_{\mathrm{JS}}(P \| Q)=\frac{1}{2} D_{\mathrm{KL}}(P \| M)+\frac{1}{2} D_{\mathrm{KL}}(Q \| M)
\end{equation}
where $M=\frac{1}{2}(P+Q)$. Finally, the probability loyalty between 
$P$ and $Q$ is defined as:
\begin{equation}
L_{p}(P \| Q)=1-\sqrt{D_{\mathrm{JS}}(P \| Q)}
\end{equation}
where $L_{p} \in [0, 1]$; higher $L_{p}$ represents higher resemblance. Note that Equation \ref{equ:kl} is also known as the KD loss~\cite{kd}, thus KD-based methods will naturally have an advantage in terms of probability loyalty.

\subsection{Robustness}
Deep Learning models 
have been shown to
be vulnerable to adversarial examples that are slightly altered with perturbations often indistinguishable to humans~\cite{kurakin2017adversarial}. Previous work~\cite{su2018robustness} found that small convolutional neural networks (CNN) are more vulnerable to adversarial attacks compared to bigger ones. 
Likewise, we intend to investigate how BERT models perform and the effect of different types of compression in terms of robustness.
We use an off-the-shelf adversarial attack method, TextFooler~\cite{jin2019bert}, which demonstrates state-of-the-art performance on attacking BERT. TextFooler conducts black-box attacks by querying the BERT model with the adversarial input where words are perturbed based on
their
part-of-speech role. We select two metrics from \cite{jin2019bert}, after-attack accuracy and query number, to evaluate a model's robustness. After-attack accuracy represents the remaining accuracy after the adversarial attack. Query number represents how many queries with perturbed input have been made to complete the attack.

\section{Experiments}

\subsection{Dataset and Baselines}
We use the Multi-Genre Natural Language Inference dataset (MNLI)~\cite{mnli} for evaluation. The accuracy scores are from the GLUE benchmark~\cite{glue} test server.
We select representative methods of different types of compression methods: Head Prune~\cite{headprune} for pruning; Post-training Quantization (PTQ) and Quantization-aware Training (QAT)~\cite{q8bert} for quantization; DistilBERT~\cite{distilbert} and TinyBERT~\cite{tinybert} for pretraining-phase knowledge distillation; BERT-PKD~\cite{bertpkd} for downstream knowledge distillation; and BERT-of-Theseus~\cite{bot} for module replacing. Following \citet{bertpkd,bot}, we truncate the first (bottom) 6 layers and then finetune it as a baseline for 6-layer models. Additionally, we directly optimize the KL divergence (\ie pure KD loss) to set an upper bound for probability loyalty.

\subsection{Training Details}
Our implementation is based on Hugging Face Transformers~\cite{transformers}.
We first finetune a BERT-base model to be the teacher for KD and the source model for quantization and pruning. The learning rate is set to $3\times10^{-5}$ and the batch size is $64$ with 1,000 warm-up steps. 
For quantization and pruning, the source model is the same finetuned teacher. For downstream KD and BERT-of-Theseus, we initialize the model by truncating the first (bottom) 6 layers of the finetuned teacher, following the original papers~\cite{bertpkd,bot}. QAT uses pretrained BERT-base for initialization. For pretraining distillation, we directly finetune compressed 6-layer DistilBERT and TinyBERT checkpoints to report results. The pruning percentage for Head Prune is $45\%$. The hyperparameters of BERT-PKD are from the original implementation. The detailed hyperparameters for each method can be found in Appendix \ref{sec:hyper}.

\subsection{Experimental Results}
\label{sec:results}
We show experimental results in Table \ref{tab:main}. First, we find that post-training quantization can drastically improve 
model
robustness.
A possible explanation is that the regularization effect of post-training quantization~\cite{paupamah2020quantisation,wu2020vector} helps improve the robustness of the model~\cite{detectingoverfitting,ma2020adversarial}. A similar but smaller effect can be found from pruning. However, as shown in Table \ref{tab:prune}, if we finetune the low-precision or pruned model again, the model would re-overfit the data and yield even lower robustness than the original model. Second, KD-based models maintains good label loyalty and probability loyalty due to their optimization objectives. Interestingly, compared to \textit{Pure KD} where we directly optimize the KL divergence, DistilBERT, TinyBERT and BERT-PKD trade some loyalty in exchange for accuracy. Compared to DistilBERT, TinyBERT digs up higher accuracy by introducing layer-to-layer distillation, with their loyalty remains identical. Also, we do not observe a significant difference between pretraining KD and downstream KD in terms of both loyalty and robustness ($p>0.1$).
Notably, BERT-of-Theseus has a significantly lower loyalty, suggesting the mechanism behind it is different from KD. We also provide some results on SST-2~\citep{sst} in Appendix \ref{sec:sst}.

\begin{table}[tb]
\begin{center}
\resizebox{1.\columnwidth}{!}{
\begin{tabular}{l|c|c|cc|rc}
\toprule
Method & Speed & MNLI & L-L & P-L & AA & \#~Q \\
\midrule
Teacher & 1.0$\times$ & $84.5$~/~$83.3$ & $100$ & $100$ & $8.1$ & $89.6$ \\
\midrule
Head Prune & 1.2$\times$ & $80.9$~/~$80.6$ & $87.8$ & $85.5$ & $9.1$ & $90.5$ \\
\xspace+Finetune & 1.2$\times$ & $83.2$~/~$81.9$ & $89.1$ & $85.5$ & $7.2$ & $83.2$ \\
\xspace+KD & 1.2$\times$ & $\mathbf{84.2}$~/~$\mathbf{83.0}$ & $\mathbf{93.3}$ & $\mathbf{93.0}$ & $8.3$ & $90.5$ \\
\xspace+KD+PTQ & 2.2$\times$ & $80.8$~/~$80.4$ & $89.6$ & $86.3$ & $\mathbf{38.4}$ & $\mathbf{90.9}$\\
\midrule
Q8-QAT & 1.8$\times$ & $83.4$~/~$82.4$ & $89.7$ & $88.2$ & $6.8$ & $82.7$\\
Q8-PTQ & 1.8$\times$ & $80.7$~/~$80.4$ & $89.6$ & $80.8$ & $\mathbf{40.2}$ & $\mathbf{91.6}$\\
\xspace+Finetune & 1.8$\times$ & $82.9$~/~$81.9$ & $89.7$ & $84.8$ & $7.1$ & $84.5$ \\
\xspace+KD & 1.8$\times$ & $\mathbf{84.1}$~/~$\mathbf{83.5}$ & $\mathbf{94.0}$ & $\mathbf{93.9}$ & $7.5$ & $86.1$ \\
\midrule
BERT-PKD & 2.0$\times$ & $81.3$~/~$81.1$ & $88.9$ & $89.0$ & $6.4$ & $81.9$ \\
Theseus & 2.0$\times$ & $81.8$~/~$80.7$ & $88.1$ & $82.5$ & $8.3$ & $89.7$ \\
\xspace+KD & 2.0$\times$ & $\mathbf{82.6}$~/~$\mathbf{81.7}$ & $\mathbf{91.2}$ & $\mathbf{91.4}$ & $8.0$ & $88.7$\\
\xspace+KD+PTQ & 3.6$\times$ & $80.2$~/~$79.9$ & $89.5$ & $80.3$ & $\mathbf{36.5}$ & $\mathbf{91.3}$\\

\bottomrule
\end{tabular}
}
\end{center}
\caption{Accuracy and loyalty for combining multiple compression techniques on the test set of MNLI. \textbf{L-L}: label loyalty; \textbf{P-L}: probability loyalty; \textbf{AA}: after-attack accuracy; \textbf{\#~Q}: Query number for adversarial attack. The number of layers for each group is consistent with Table \ref{tab:main}.}
\label{tab:prune}
\end{table}

\section{Combining the Bag of Tricks}
As we described in Section \ref{sec:results}, we discover that post-training quantization (PTQ) can improve the robustness of a model while knowledge distillation (KD) loss benefits the loyalty of a compressed model. Thus, by combining multiple compression techniques, we expect to achieve a higher speed-up ratio with improved accuracy, loyalty and robustness. 

To combine KD with other methods, we replace the original cross-entropy loss in quantization-aware training and module replacing with the knowledge distillation loss~\cite{kd} as in Equation \ref{equ:kl}. For pruning, we perform knowledge distillation on the pruned model.
We also apply the temperature re-scaling trick from~\cite{kd} with a fixed temperature 
of
10. As shown in Table \ref{tab:prune}, the knowledge distillation loss effectively improves the accuracy
and loyalty of pruning, quantization and module replacing.

Furthermore, we post-quantize the KD-enhanced models after they are trained. Shown in Table \ref{tab:prune}, by adding post-training quantization, the speed and robustness can
both be boosted. Notably, the order to apply PTQ and KD does matter. PTQ$\rightarrow$KD has high accuracy and loyalty but poor robustness while KD$\rightarrow$PTQ remains a good robustness with a lower accuracy performance.
To summarize, we recommend the following compression strategy: (1) conduct pruning or module replacing with a KD loss; (2) for speed-sensitive and robustness-sensitive applications, apply post-training quantization afterwards.

\section{Conclusion}
In this paper, we propose label and probability loyalty to measure the correspondence of label and predicted probability distribution between
compressed and original models. In addition to loyalty, we investigate the robustness of different compression techniques under adversarial attacks. These metrics reveal that post-training quantization and knowledge distillation can drastically improve 
robustness and loyalty, respectively. By combining multiple compression methods, we can further improve 
speed, accuracy, loyalty and robustness for various applications. Our metrics help mitigate the gap between model training and deployment, shed light upon comprehensive evaluation for compression of pretrained language models, and call for the invention of new compression techniques.

\section*{Ethical Concerns}
We include a discussion about the possible ethical risks of a compressed model in Section 3.1. Although our paper is an attempt to mitigate the risk of introducing extra biases to compression, we would like to point out that our metrics do not directly indicate the bias level in the compressed model. That is to say, additional measures should still be taken to evaluate and debias both the teacher and student models.

\section*{Acknowledgments}
We would like to thank all anonymous reviewers for their insightful comments. Tao Ge is the corresponding author.

\bibliographystyle{acl_natbib}
\bibliography{custom}

\newpage
\appendix

\section{Hyperparameter Settings}
\label{sec:hyper}
\paragraph{Teacher}
Learning rate: 3e-5; Batch size: 64; Warm-up steps: 1000.
\paragraph{Truncate \& Finetune}
Learning rate: 3e-5; Batch size: 128.
\paragraph{Pure KD}
Learning rate: 3e-5; Batch size: 64; Warm-up steps: 1000; $\alpha$: 1; Temperature: 10.
\paragraph{Q8-QAT}
Learning rate: 2e-5; Batch size: 128.
\paragraph{DistilBERT}
Learning rate: 3e-5; Batch size: 64; Warm-up steps: 1000.
\paragraph{TinyBERT}
Learning rate: 3e-5; Batch size: 64; Warm-up steps: 1000.
\paragraph{BERT-PKD}
Learning rate: 3e-5; Batch size: 64; Warm-up steps: 1000; $\alpha$: 0.7; $\beta$: 500; Temperature: 10.
\paragraph{BERT-of-Theseus}
Learning rate: 3e-5; Batch size: 128; Warm-up steps: 8000; $k$: 0.00002; $b$: 0.5.
\paragraph{Head Prune + Finetune}
Learning rate: 2e-5; Batch size: 128.
\paragraph{Head Prune + KD}
Learning rate: 2e-5; Batch size: 128; Temperature: 10.
\paragraph{Q8 + FT}
Learning rate: 2e-5; Batch size: 128.
\paragraph{Q8 + KD}
Learning rate: 2e-5; Batch size: 128; Temperature: 10.
\paragraph{BERT-of-Theseus + KD}
Learning rate: 3e-5; Batch size: 64; Warm-up steps: 1000; Temperature: 10.

\section{Experimental Results on SST-2}
\label{sec:sst}

\begin{table}[h]
\begin{center}
\resizebox{1.\columnwidth}{!}{
\begin{tabular}{l|c|c|cc|rc}
\toprule
Method & Speed & SST-2 & L-L & P-L & AA & \#~Q \\
\midrule
Teacher & 1.0$\times$ & $92.0$ & $100$ & $100$ & $7.5$ & $81.2$ \\
\midrule
KD & 2.0$\times$ & $\mathbf{91.5}$ & $\mathbf{93.8}$ & $\mathbf{92.4}$ & $7.2$ & $80.3$ \\
\midrule
Head Prune & 1.3$\times$ & $90.4$ & $89.5$ & $88.2$ & $8.1$ & $81.0$ \\
\midrule
Q8-QAT & 1.8$\times$ & $91.4$ & $91.8$ & $90.3$ & $6.6$ & $81.4$\\
Q8-PTQ & 1.8$\times$ & $90.1$ & $91.3$ & $88.9$ & $\mathbf{26.5}$ & $\mathbf{86.4}$\\

\bottomrule
\end{tabular}
}
\end{center}
\caption{Accuracy and loyalty of some compression techniques on the test set of SST-2. \textbf{L-L}: label loyalty; \textbf{P-L}: probability loyalty; \textbf{AA}: after-attack accuracy; \textbf{\#~Q}: Query number for adversarial attack. The number of layers for each group is consistent with Table \ref{tab:main}.}
\label{tab:sst2}
\end{table}

\end{document}